\documentclass[10pt,journal,compsoc]{IEEEtran}
\ifCLASSOPTIONcompsoc
  \usepackage[nocompress]{cite}
\else
  \usepackage{cite}
\fi
\usepackage{graphicx}
\ifCLASSINFOpdf
\else
\fi
\usepackage{url}

\usepackage{xcolor}

\usepackage{multirow}

\begin{document}
%
\title{Show and Tell: Lessons learned from the 2015 MSCOCO Image Captioning Challenge}
%
%
%
%

\author{Oriol~Vinyals,
        Alexander~Toshev,
        Samy~Bengio,
        and~Dumitru~Erhan}
\markboth{IEEE Transaction on Pattern Analysis and Machine Intelligence,~Vol.XX, No.~XX, Month~2016}%
{Shell \MakeLowercase{\textit{et al.}}: Bare Demo of IEEEtran.cls for Computer Society Journals}
%

\IEEEpubid{Copyright (c) 2016 IEEE. Personal use is permitted. For any other purposes, permission must be obtained from the IEEE. DOI:  \url{http://dx.doi.org/10.1109/TPAMI.2016.2587640}}


\IEEEtitleabstractindextext{%
\begin{abstract}
Automatically describing the content of an image is a fundamental
problem in artificial intelligence that connects
computer vision and natural language processing.
In this paper, we present a generative model based on a deep recurrent
architecture that combines recent advances in computer vision and
machine translation and that can be used to generate natural sentences
describing an image.  The model is trained
to maximize the likelihood of the target description
sentence given the training image.  Experiments on several datasets show
the accuracy of the model and the fluency of the language it learns
solely from image descriptions. Our model is often quite accurate,
which we verify both qualitatively and quantitatively.
Finally, given the recent surge of interest in this task, a competition
was organized in 2015 using the newly released COCO dataset. We describe
and analyze the various improvements we applied to our own baseline and
show the resulting performance in the competition, which we won {\em ex-aequo}
with a team from Microsoft Research, and provide an open source implementation in TensorFlow.
\end{abstract}


}

\maketitle

\IEEEdisplaynontitleabstractindextext

%
\IEEEpeerreviewmaketitle

\IEEEraisesectionheading{\section{Introduction}\label{sec:intro}}

\IEEEPARstart{B}{eing}
able to automatically describe the content of an image using properly
formed English sentences is a very challenging task, but it could have great
impact, for instance by helping visually impaired people better understand the
content of images on the web. This task is significantly harder, for example, than the
well-studied image classification or object recognition tasks,
which have been a main focus in the computer vision community~\cite{ILSVRCarxiv14}.
Indeed, a description must capture not only the objects contained in an image, but
it also must express how these objects relate to each other as
well as their attributes and the activities they are involved in. Moreover, the above
semantic knowledge has to be expressed in a natural language like English, which
means that a language model is needed in addition to visual understanding.

Most previous attempts have proposed
to stitch together existing solutions of the above sub-problems, in order to go from
an image to its description~\cite{farhadi2010every,kulkarni2011baby}. In contrast, we would like to 
present in this work a single joint model that
takes an image $I$ as input, and is trained to maximize the likelihood
$p(S|I)$ of producing a target sequence of words $S = \{S_1, S_2, \ldots\}$
where each word $S_t$ comes from a given dictionary, that describes the image
adequately.

\begin{figure}
\begin{center}
\includegraphics[width=0.5\textwidth]{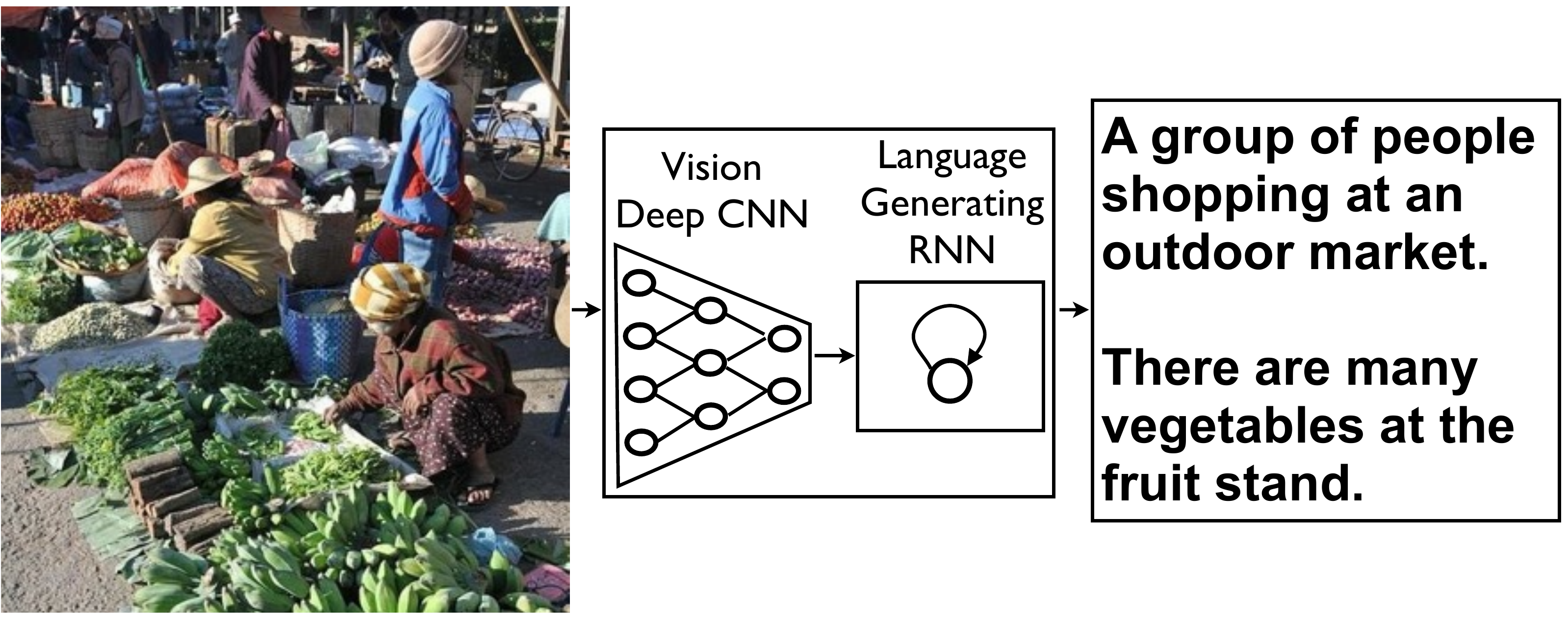}
\end{center}
\caption{\label{fig:overview} NIC, our model, is based end-to-end on a neural network consisting of a vision CNN followed by a language generating RNN. It generates complete sentences in natural language from an input image, as shown on the example above.}
\end{figure}

The main inspiration of our work comes from recent advances in machine translation, where the task is to transform a sentence $S$ written
in a source language, into its translation $T$ in the target language, by
maximizing $p(T|S)$. For many
years, machine translation was also achieved by a series of separate tasks
(translating words individually, aligning words, reordering, etc), but recent
work has shown that translation can be done in a much simpler way using 
Recurrent Neural Networks
(RNNs)~\cite{cho2014learning,bahdanau2014neural,sutskever2014sequence}
and still reach state-of-the-art performance.
An ``encoder'' RNN {\em reads} the source sentence and
transforms it into a rich fixed-length vector representation, which in turn in used as the 
initial hidden state of a ``decoder'' RNN that {\em generates}
the target sentence.

Here, we propose to follow this elegant recipe,
replacing the encoder RNN by a deep convolution neural network (CNN). Over the last few years it has been convincingly 
shown that CNNs can produce a rich representation of the input image by embedding it
to a fixed-length vector, such that this representation can be used for a variety of
vision tasks~\cite{sermanet2013overfeat}. Hence, it is natural to use a CNN as an
image ``encoder'', by first pre-training it for an image classification task and
using the last hidden layer as an input to the RNN decoder that generates sentences (see Fig.~\ref{fig:overview}).
We call this model the Neural Image Caption, or NIC.

Our contributions are as follows. First, we present an end-to-end system for the
problem. It is a neural net which is fully trainable using stochastic
gradient descent.
Second, our model combines state-of-art sub-networks for vision and language models. These
can be pre-trained on larger corpora and thus can take advantage of additional data. Finally,  
it yields significantly better performance compared to state-of-the-art approaches;
for instance, on the Pascal dataset, NIC yielded a BLEU score of 59,
to be compared to the current state-of-the-art of 25, while human performance
reaches 69. On Flickr30k, we improve from 56 to 66, and on SBU,
from 19 to 28. Third, we describe the lessons learned from participating in the first MSCOCO
competition, which helped us to improve our initial model and place first in automatic metrics, and first
(tied with another team) in human evaluation.

\section{Related Work}
\label{sec:related}

The problem of generating natural language descriptions from visual
data has long been studied in computer vision, but mainly for
video~\cite{gerber1996knowledge,yao2010i2t}. Traditionally, this has led to complex
systems composed of visual primitive recognizers combined with a structured
formal language, e.g.~And-Or Graphs or logic systems, which are
further converted to natural language via rule-based systems. Such
systems are heavily hand-designed, relatively brittle and have been
demonstrated only on limited domains, e.g. traffic scenes or sports.

The problem of still image captioning in natural language has recently enjoyed 
increased interest. Recent advances in object recognition and detection
as well as attribute recognition has been used to drive natural language
generation systems, though these are limited in their
expressivity. Farhadi et al.~\cite{farhadi2010every} use detections to
infer a triplet of scene elements which is converted to text using
templates. Similarly, Li et al.~\cite{li2011composing} start off with
detections and piece together a final description using phrases containing
detected objects and relationships. A more complex graph of detections
beyond triplets is used by Kulkani et
al.~\cite{kulkarni2011baby}, but with template-based text generation.
More powerful language models based on language parsing
have been used as well
\cite{mitchell2012midge,aker2010generating,kuznetsova2012collective,kuznetsova2014treetalk,elliott2013image}. The
above approaches have been able to describe images ``in the wild",
but they are heavily hand-designed and rigid when it comes to text
generation.

A large body of work has addressed the problem of ranking descriptions
for a given image
\cite{hodosh2013framing,gong2014improving,ordonez2011im2text,devlin2015exploring,kolavr2015technical}. Such
approaches are based on the idea of co-embedding of images and text in
the same vector space. For an image query, descriptions are retrieved
which lie close to the image in the embedding space. Most closely, neural networks are used to co-embed
images and sentences together \cite{socher2014grounded} or even image crops and subsentences \cite{karpathy2014deep} but do not attempt to generate novel
descriptions. In general, the above approaches cannot describe previously unseen
compositions of objects, even though the individual objects might have been
observed in the training data. Moreover, they avoid addressing the
problem of evaluating how good a generated description is. More recently neural net based recognizer are used to detect a larger set of words and in conjunction with a language model sentences are generated \cite{msr14}. 

In this work we combine deep
convolutional nets for image classification \cite{batchnorm} with
recurrent networks for sequence modeling
\cite{hochreiter1997long}, to create a single network
that generates descriptions of images. The RNN is trained in the context of
this single ``end-to-end'' network. The model is inspired
by recent successes of sequence generation in machine translation
\cite{cho2014learning,bahdanau2014neural,sutskever2014sequence}, with
the difference that instead of starting with a sentence, we provide an image
processed by a convolutional net. 

In the summer of 2015 a few approaches were introduced which follow the above general paradigm. The closest works are by Kiros et al.~\cite{kiros2013multimodal} who
use a neural net, but a feedforward one, to predict the next word given the image
and previous words. A recent work by Mao et al.~\cite{baidu2014,mao2014deep} uses a recurrent
NN for the same prediction task. This is very similar to the present proposal but
there are a number of important differences:  we use a more powerful RNN model,
and provide the visual input to the RNN model directly, which makes it possible
for the RNN to keep track of the objects that have been explained by the text.  As
a result of these seemingly insignificant differences, our system achieves
substantially better results on the established benchmarks. Further, Kiros et al.~\cite{kiros2014}
propose to construct a joint multimodal embedding space by using a powerful
computer vision model and an LSTM that encodes text. In contrast to our approach,
they use two separate pathways (one for images, one for text) to define a joint embedding,
and, even though they can generate text, their approach is highly tuned for ranking. A recurrent network is being used by Donahue et al.~\cite{berkeley2014} who address in addition activity recognition and video description.

In addition, some approaches try to model in a more explicit fashion the visual anchoring of sentence parts claiming a performance benefit. Xu et al.~\cite{xu2015show} explore attention mechanisms over image regions where while emitting words the system can focus on image parts. An explicit word to region alignment is utilized during training by Karpathy et al.~\cite{karpathy14}. Finally, Chen et al.~\cite{chen1997mind} build a visual representation for sentence parts while generating the description. Further analysis of the above approaches were reported by Devlin et al.~\cite{devlin2015language}.

\section{Model}
\label{sec:model}

In this paper, we propose a neural and probabilistic framework to generate
descriptions from images. Recent advances in statistical machine
translation have shown that, given a powerful sequence model, it is
possible to achieve state-of-the-art results by directly maximizing
the probability of the correct translation given an input sentence in
an ``end-to-end'' fashion -- both for training and inference. These
models make use of a recurrent neural network
which encodes the variable length input into a fixed dimensional
vector, and uses this representation to ``decode'' it to the desired
output sentence. Thus, it is natural to use the same approach where,
given an image (instead of an input sentence in the source language),
one applies the same principle of ``translating'' it into its
description.

Thus, we propose to directly maximize the probability of the correct
description given the image by using the following formulation:

\begin{equation}
\theta^\star = \arg\max_\theta \sum_{(I,S)} \log p(S | I ; \theta)
\label{eqn:obj}
\end{equation}
where $\theta$ are the parameters of our model, $I$ is an image, and
$S$ its correct transcription. Since $S$ represents any sentence, its
length is unbounded. Thus, it is common to apply the chain rule to
model the joint probability over $S_0,\ldots,S_N$, where $N$ is the
length of this particular example as

\begin{equation}
\log p(S | I) = \sum_{t=0}^N \log p(S_t | I, S_0, \ldots, S_{t-1})
\label{eqn:chain}
\end{equation}
where we dropped the dependency on $\theta$ for convenience.
At training
time, $(S,I)$ is a training example pair, and we optimize the sum of
the log probabilities as described in~(\ref{eqn:chain}) over the
whole training set using stochastic gradient descent (further training
details are given in Section \ref{sec:exps}).

It is natural to model $p(S_t | I, S_0, \ldots, S_{t-1})$ with a
Recurrent Neural Network (RNN), where the variable number of 
words we condition upon up to $t-1$ is expressed by a fixed length 
hidden state or memory $h_t$. This memory is updated after seeing a
new input $x_t$  by using a non-linear function $f$:
\begin{equation}\label{eq:rnn}
h_{t+1} = f(h_{t}, x_t)\;.
\end{equation}
To make the above RNN more concrete two crucial design choices are to be made: what is
the exact form of $f$ and how are the images and words fed as inputs $x_t$. For 
$f$ we use a Long-Short Term Memory (LSTM) net, which has shown state-of-the art
performance on sequence tasks such as translation. This model is outlined in the
next section. 

For the representation of images, we use a Convolutional Neural Network
(CNN). They have been widely used and studied for image tasks, and are
currently state-of-the art for object recognition and detection. Our particular
choice of CNN uses the recent approach of batch normalization and yields the
current best performance on the ILSVRC 2014 classification
competition~\cite{batchnorm}. Furthermore, they have been shown to
generalize to other tasks such as scene classification by means of
transfer learning~\cite{decaf2014}. The words are represented with an embedding
model~\cite{mikolov2013}.

\subsection{LSTM-based Sentence Generator}
\label{sec:lstm}

The choice of $f$ in (\ref{eq:rnn}) is governed by its 
ability to deal with vanishing and exploding gradients~\cite{hochreiter1997long},
the most common
challenge in designing and training RNNs. To address this challenge, a  particular form 
of recurrent nets, called LSTM, was introduced \cite{hochreiter1997long}
and applied with great success to translation \cite{cho2014learning,sutskever2014sequence} and sequence generation \cite{graves2013generating}.

\begin{figure}
\begin{center}
\includegraphics[width=0.85\columnwidth]{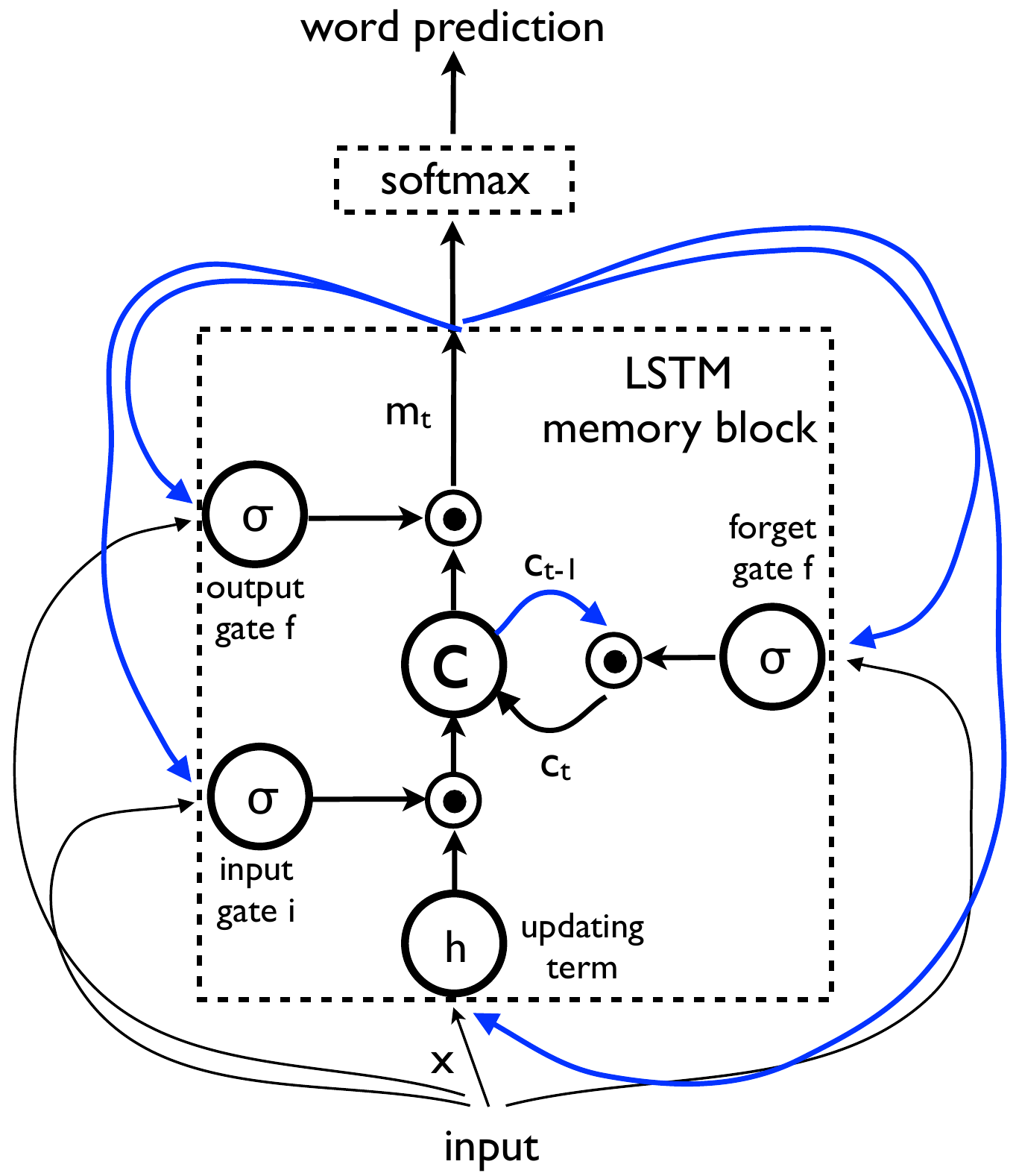}
\end{center}
\caption{\label{fig:lstm} LSTM: the memory block contains a cell $c$ which is controlled by three gates. In blue we show the recurrent connections -- the output $m$ at time $t-1$ is fed back to the memory at time $t$ via the three gates; the cell value is fed back via the forget gate; the predicted word at time $t-1$ is fed back in addition to the memory output $m$ at time $t$ into the Softmax for word prediction.}
\end{figure}

The core of the LSTM model is a memory cell $c$ encoding
knowledge at every time step of what inputs have been observed up to this step (see Figure~\ref{fig:lstm}) . The behavior of the cell
is controlled by ``gates" -- layers which are applied multiplicatively and thus can
either keep a value from the gated layer if the gate is $1$ or zero this value if the gate is $0$.
In particular, three gates are being used which control whether to forget the current cell value (forget gate $f$),
if it should read its input (input gate $i$) and whether to output the new cell value (output gate $o$).
The definition of the gates and cell update and output are as follows:
\begin{eqnarray}
i_t &= &\sigma(W_{ix} x_t+ W_{im} m_{t-1}) \\
f_t  &= & \sigma(W_{fx} x_t+ W_{fm} m_{t-1}) \\
o_t  &= & \sigma(W_{ox} x_t + W_{om} m_{t-1})  \\
c_t  &= & f_t \odot c_{t-1} + i_t \odot h(W_{cx} x_t + W_{cm} m_{t-1}) \\
m_t  &= & o_t \odot c_t \\
p_{t+1} &=& \textrm{Softmax}(m_t) 
\end{eqnarray}
where $\odot$ represents the product with a gate value, and the various $W$
matrices are trained parameters. Such multiplicative gates make it
possible to train the LSTM robustly as these gates deal well with exploding and vanishing gradients \cite{hochreiter1997long}.
The nonlinearities are sigmoid $\sigma(\cdot)$ and hyperbolic tangent $h(\cdot)$.
The last equation $m_t$ is what is used
to feed to a Softmax, which will produce a probability distribution $p_t$ over all words.

\begin{figure}
\begin{center}
\includegraphics[width=0.75\columnwidth]{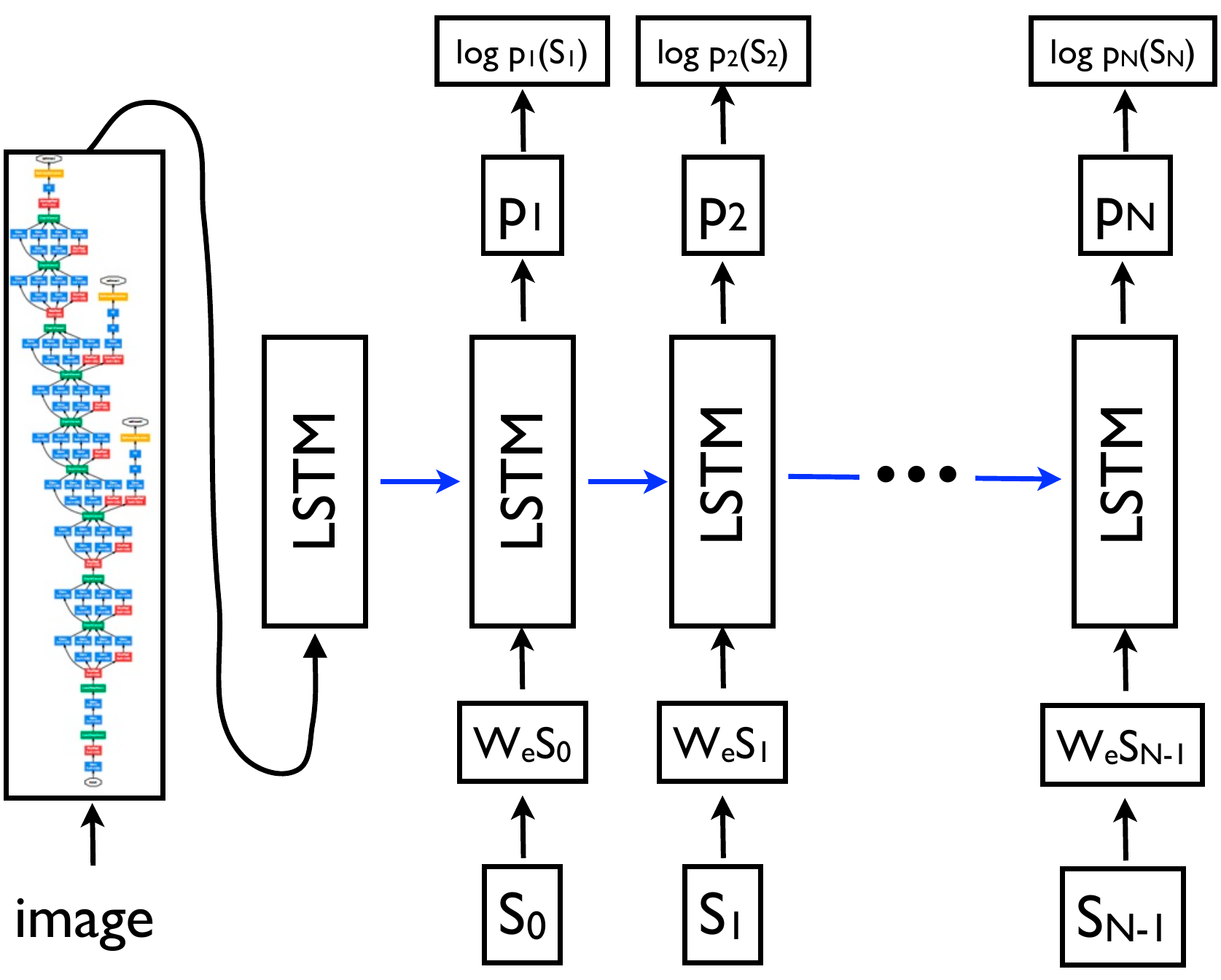}
\end{center}
\caption{\label{fig:unrolled_lstm} LSTM model combined with a CNN image embedder (as defined in \cite{batchnorm}) and word embeddings. The unrolled connections between the LSTM memories are in blue and they correspond to the recurrent connections in Figure~\ref{fig:lstm}. All LSTMs share the same parameters. }
\end{figure}
\subsubsection{Training} The LSTM model is trained to predict each word of the
sentence after it has seen the image as well as all preceding words as defined by
$p(S_t | I, S_0, \ldots, S_{t-1})$. For this purpose, it is instructive to think
of the LSTM in unrolled form -- a copy of the LSTM memory is created for the
image and each sentence word such that all LSTMs share the same parameters and the
output $m_{t-1}$ of the LSTM at time $t-1$ is fed to the LSTM at time $t$ (see
Figure~\ref{fig:unrolled_lstm}). All recurrent connections are transformed to feed-forward connections in the 
unrolled version. In more detail, if we denote by $I$ the input
image and by $S=(S_0,\ldots, S_N)$ a true sentence describing this image, the
unrolling procedure reads:
\begin{eqnarray}
x_{-1} &=& \textrm{CNN}(I)\\
x_t &=& W_e S_t, \quad t\in\{0\ldots N-1\}\quad \label{eqn:sparse}\\
p_{t+1} &=& \textrm{LSTM}(x_t), \quad t\in\{0\ldots N-1\}\quad
\end{eqnarray}
where we represent each word as a one-hot vector $S_t$ of dimension equal to the
size of the dictionary. Note that we denote by $S_0$ a special start word and by
$S_{N}$ a special stop word which designates the start and end of the sentence.
In particular by emitting the stop word the LSTM signals that a complete sentence
has been generated. Both the image and the words are mapped to the same space,
the image by using a vision CNN, the words by using word embedding $W_e$. The image
$I$ is only input once, at $t=-1$, to inform the LSTM about the image contents. We
empirically verified that feeding the image at each time step as an extra input yields
inferior results, as the network can explicitly exploit noise in the image and
overfits more easily.

Our loss is the sum of the negative log likelihood of the correct word at each step as follows:
\begin{equation}
L(I, S) = - \sum_{t=1}^N \log p_t(S_t) \; .
\end{equation}
The above loss is minimized w.r.t. all the parameters of the LSTM, the top layer of the
image embedder CNN and word embeddings $W_e$.

\subsubsection{Inference}

There are multiple approaches that can be used to generate a sentence given
an image, with NIC. The first one is {\bf Sampling} where we just
sample the first word according to $p_1$, then provide the corresponding
embedding as input and sample $p_2$, continuing like this until we sample the
special end-of-sentence token or some maximum length.
The second  one is {\bf BeamSearch}: iteratively
consider the set of the $k$ best sentences up to time
$t$ as candidates to generate sentences of size $t+1$, and keep only the
resulting best $k$ of them. This better approximates
$S = \arg\max_{S'} p(S'|I)$.
We used the BeamSearch approach in the following experiments, with a
beam of size 20. Using a beam size of 1 (i.e., greedy search) did degrade our
results by 2 BLEU points on average.
Further experiments on varying the beam size are reported in
Section~\ref{sec:beam_size_reduction}.

\section{Experiments}
\label{sec:exps}
We performed an extensive set of experiments to assess the effectiveness of our
model using several metrics, data sources, and model architectures, in order
to compare to prior art.

\subsection{Evaluation Metrics}
Although it is sometimes not clear whether a description should be deemed
successful or not given an image,
prior art has proposed several evaluation metrics. The most
reliable (but time consuming) is to ask for raters to give a subjective score
on the usefulness of each description given the image. In this paper, we used
this to reinforce that some of the automatic metrics indeed correlate with this
subjective score, following the guidelines proposed
in~\cite{hodosh2013framing}, which asks the
graders to evaluate each generated sentence with a scale from 1 to 4\footnote{
The raters are asked whether the image is
described without any errors, described with minor errors, with a somewhat
related description, or with an unrelated description, with a score of 4 being
the best and 1 being the worst.}.

For this metric, we set up an Amazon Mechanical Turk experiment. Each image was
rated by 2 workers. The typical level of agreement between workers
is $65\%$. In case of disagreement we simply average the scores and record the
average as the score. For variance analysis, we perform bootstrapping
(re-sampling the results with replacement and computing means/standard
deviation over the resampled results). Like~\cite{hodosh2013framing} we
report the fraction
of scores which are larger or equal than a set of predefined thresholds.

The rest of the metrics can be computed automatically assuming one has access to
groundtruth, i.e.~human generated descriptions. The most commonly used metric
so far in the image description literature has been the
BLEU score~\cite{papineni2002},
which is a form of precision of word n-grams between generated and reference
sentences~\footnote{In this literature, most previous work report BLEU-1, i.e., they only compute precision at the unigram level, whereas BLEU-n is a geometric average of precision over 1- to n-grams.}.
Even though this metric has some obvious drawbacks, it has been shown to correlate
well with human evaluations. In this work, we corroborate this as well, as
we show in Section~\ref{sec:results}. An extensive evaluation protocol, as well
as the generated outputs of our system, can be found at \url{http://nic.droppages.com/}.

Besides BLEU, one can use the perplexity of the model for a given transcription
(which is closely related to our objective function in (\ref{eqn:obj})). The perplexity
is the geometric mean of the inverse probability for each predicted word. We
used this metric to perform choices regarding model selection and hyperparameter
tuning in our held-out set, but we do not report it since BLEU is always preferred
\footnote{Even though it would be more desirable, optimizing for BLEU score yields
a discrete optimization problem. In general, perplexity and BLEU scores are fairly
correlated.}. 

More recently, a novel metric called CIDER \cite{cider} has been introduced and used by the 
organizers of the MS COCO Captioning challenge. In a nutshell, it measures consistency 
between n-gram occurrences in generated and reference sentences, where this consistency
is weighted by n-gram saliency and rarity.

As all of the above metrics have various shortcomings (see \cite{cider} for detailed discussion), we 
provide in addition results using METEOR~\cite{banerjee2005meteor} and ROUGE~\cite{lin2004rouge} metrics.

Lastly, the current literature on image description
has also been using the proxy task of ranking a set of available
descriptions with respect to a given image (see for instance~\cite{kiros2014}).
Doing so has the advantage that one can use known ranking metrics like recall@k.
On the other hand, transforming the description generation task into a ranking
task is unsatisfactory: as the complexity of images to describe grows, together
with its dictionary, the number of possible sentences grows exponentially with
the size of the dictionary, and
the likelihood that a predefined sentence will fit a new image will go down
unless the number of such sentences also grows exponentially, which is not
realistic; not to mention the underlying computational complexity of evaluating
efficiently such a large corpus of stored sentences for each image.
The same argument has been used in speech recognition, where one has to
produce the sentence corresponding to a given acoustic sequence; while early
attempts concentrated on classification of isolated phonemes or words,
state-of-the-art approaches for this task are now generative and can produce
sentences from a large dictionary.

Now that our models can generate descriptions of reasonable quality,
and despite the ambiguities of evaluating an image description (where there
could be multiple valid descriptions not in the groundtruth)
we believe we should concentrate on evaluation metrics for the generation task
rather than for ranking.

\subsection{Datasets}
\label{sec:data}
For evaluation we use a number of datasets which consist of images and sentences in English describing these
images. The statistics of the datasets are as follows:
\begin{center}
\begin{tabular}{|l|c|c|c|}
\hline
\multirow{2}{*}{Dataset name} & \multicolumn{3}{|c|}{size} \\
\cline{2-4}
 & train & valid. & test \\
\hline
\hline
Pascal VOC 2008 \cite{farhadi2010every} & - & - & 1000 \\
\hline
Flickr8k \cite{rashtchian2010collecting} & 6000 & 1000 & 1000 \\
\hline
Flickr30k \cite{hodoshimage} & 28000 & 1000 & 1000 \\
\hline
MSCOCO \cite{lin2014microsoft} & 82783 & 40504 & 40775 \\
\hline
SBU \cite{ordonez2011im2text} & 1M & - & - \\
\hline
\end{tabular}
\end{center}
With the exception of SBU, each image has been annotated by labelers
with 5 sentences that are
relatively visual and unbiased. SBU consists of
descriptions given by image owners when they uploaded them to Flickr. As 
such they are not guaranteed to be visual or unbiased and thus this dataset has more noise.

The Pascal dataset is customary used for testing only after a system has been trained on 
different data such as any of the other four dataset. In the case of SBU, we hold
out 1000 images for testing and train on the rest as 
used by \cite{kuznetsova2014treetalk}. Similarly, we reserve 4K random images from the
MSCOCO validation set as test, called COCO-4k, and use it to report results in the following section.

\subsection{Results}
\label{sec:results}

Since our model is data driven and trained end-to-end, and given the abundance of
datasets, we wanted to answer
questions such as ``how dataset size affects generalization'',
``what kinds of transfer learning it would be able to achieve'',
and ``how it would deal with weakly labeled examples''.
As a result, we performed experiments on five different datasets,
explained in Section~\ref{sec:data}, which enabled us to understand
our model in depth.

\subsubsection{Training Details}

Many of the challenges that we faced when training our models had to do with overfitting.
Indeed, purely supervised approaches require large amounts of data, but the datasets
that are of high quality have less than 100000 images. The task
of assigning a description is strictly harder than object classification and
data driven approaches have only recently become dominant thanks to datasets as large as ImageNet
(with ten times more data than the datasets we described in this paper, with the exception of SBU).
As a result, we believe that, even with the results we obtained which are quite good, the advantage
of our method versus most current human-engineered approaches will only increase in the next few years as training set sizes will grow.

Nonetheless, we explored several techniques to deal with overfitting. The most obvious
way to not overfit is to initialize the weights of the CNN component of our system to
a pretrained model (e.g., on ImageNet). We did this in all the experiments (similar to~\cite{gong2014improving}),
and it did help quite a lot in terms of generalization. Another set of weights that could
be sensibly initialized are $W_e$, the word embeddings. We tried initializing them
from a large news corpus~\cite{mikolov2013}, but no significant gains were observed, and we decided
to just leave them uninitialized for simplicity. Lastly, we did some model level overfitting-avoiding
techniques. We tried dropout~\cite{zaremba2014} and ensembling models, as well as exploring the size
(i.e., capacity) of the model by trading off number of hidden units versus depth. Dropout and ensembling
gave a few BLEU points improvement, and that is what we report throughout the paper. Further details 
of the ensambling and additional training improvements used for the MS COCO challenge are 
described in Section~\ref{sec:improvements}.

We trained all sets of weights using stochastic gradient descent
with fixed learning rate and no momentum.
All weights were randomly initialized except for the CNN weights,
which we left unchanged because changing them had a negative impact.
We used 512 dimensions for the embeddings and the size of the LSTM memory.

Descriptions were preprocessed with basic tokenization, keeping all words
that appeared at least 5 times in the training set.

\subsubsection{Generation Results}

We report our main results on all the relevant datasets in Tables~\ref{tab:coco} and \ref{tab:bleu}.
Since PASCAL does not have a training set, we used the system trained using MSCOCO (arguably
the largest and highest quality dataset for this task). The state-of-the-art results for PASCAL
and SBU did not use image features based on deep learning, so arguably a big improvement
on those scores comes from that change alone. The Flickr datasets have been used
recently~\cite{hodosh2013framing,baidu2014,kiros2014}, but mostly evaluated in a retrieval framework. A
notable exception is~\cite{baidu2014}, where they did both retrieval and generation, and which
yields the best performance on the Flickr datasets up to now.

Human scores in Table~\ref{tab:bleu} were computed by comparing one of the human captions against the other four.
We do this for each of the five raters, and average their BLEU scores. Since this gives a slight
advantage to our system, given the BLEU score is computed against five reference sentences
and not four, we add back to the human scores the average difference of having five references instead of four.

Given that the field has seen significant advances in the last years, we do think
it is more meaningful to report BLEU-4, which is the standard in machine translation moving forward. Additionally,
we report metrics shown to correlate better with human evaluations in Table~\ref{tab:coco}\footnote{We
used the implementation of these metrics kindly provided in \url{http://www.mscoco.org}.}.
Despite recent efforts on better evaluation metrics \cite{cider}, our model fares strongly versus
human raters. However, when evaluating our captions using human raters (see Section~\ref{sec:human}),
our model fares much more poorly, suggesting more work is needed towards better metrics.
For a more detailed description and comparison of our results on the MSCOCO dataset, and other interesting human metrics, see Section~\ref{sec:competition}. In that section, we detail the lessons learned from extra tuning of our model w.r.t. the original model which was submitted in a previous version of this manuscript \cite{google2014} (NIC in Table~\ref{tab:coco}) versus the latest version for the competition (NICv2 in Table~\ref{tab:coco}).

\begin{table}
\caption{Scores on the MSCOCO development set for two models: NIC, which was the model which we developed in \cite{google2014}, and NICv2, which was the model after we tuned and refined our system for the MSCOCO competition.}\label{tab:coco}
\centering
\begin{small}
\begin{tabular}{|c|c|c|c|}
 \hline
Metric & BLEU-4 & METEOR & CIDER \\
\hline
\hline
NIC   & 27.7  & 23.7 & 85.5     \\
NICv2   & \bf{32.1}  & \bf{25.7} & \bf{99.8}     \\
\hline
Random  &   4.6    &  9.0    &  5.1    \\
Nearest Neighbor & 9.9  & 15.7  & 36.5   \\
Human   & 21.7  & 25.2 & 85.4 \\
\hline
\end{tabular}
\end{small}
\end{table}

\begin{table}
\caption{BLEU-1 scores. We only report previous work
results when available. SOTA stands for the current
state-of-the-art.}\label{tab:bleu}
\centering
\begin{small}
\begin{tabular}{|c|c|c|c|c|}
 \hline
Approach & PASCAL & Flickr& Flickr& SBU  \\
         & (xfer) & 30k   & 8k    &      \\
\hline
\hline
Im2Text~\cite{ordonez2011im2text} &     &     &     & 11     \\
TreeTalk~\cite{kuznetsova2014treetalk} &     &     &     & 19     \\
BabyTalk~\cite{kulkarni2011baby} & 25  &     &     &        \\
Tri5Sem~\cite{hodosh2013framing} &     &     & 48  &        \\
m-RNN~\cite{baidu2014} &     & 55  & 58  &        \\
MNLM~\cite{kiros2014}\footnotemark &     & 56  & 51  &        \\
\hline
SOTA      & 25  & 56  & 58  & 19     \\
\hline
NIC   & \bf{59}  & \bf{66} & \bf{63}  & \bf{28}     \\
\hline
Human   & 69  & 68 & 70  &        \\
\hline
\end{tabular}
\end{small}
\end{table}

\footnotetext{We computed these BLEU scores with the outputs that the authors of \cite{kiros2014} kindly provided for their OxfordNet system.}

\subsubsection{Transfer Learning, Data Size and Label Quality}

Since we have trained many models and we have several testing sets, we wanted to
study whether we could transfer a model to a different dataset, and how much the
mismatch in domain would be compensated with e.g. higher quality labels or more training
data.

The most obvious case for transfer learning and data size is between Flickr30k and Flickr8k. The two
datasets are similarly labeled as they were created by the same group.
Indeed, when training on Flickr30k (with about 4 times more training data),
the results obtained are 4 BLEU points better.
It is clear that in this case, we see gains by adding more training data
since the whole process is data-driven and overfitting prone.
MSCOCO is even bigger (5 times more
training data than Flickr30k), but since the collection process was done differently, there are likely
more differences in vocabulary and a larger mismatch. Indeed, all the BLEU scores degrade by 10 points.
Nonetheless, the descriptions are still reasonable.

Since PASCAL has no official training set and was collected independently of Flickr and MSCOCO, we
report transfer learning from MSCOCO (in Table~\ref{tab:bleu}). Doing transfer learning from
Flickr30k yielded worse results with BLEU-1 at 53 (cf. 59).

Lastly, even though SBU has weak labeling (i.e., the labels were captions and not
human generated descriptions), the task is much harder with a much larger and noisier
vocabulary. However, much more data is available for training. When running the MSCOCO
model on SBU, our performance degrades from 28 down to 16.

\subsubsection{Generation Diversity Discussion}

Having trained a generative model that gives $p(S|I)$, an obvious question is
whether the model generates novel captions, and whether the generated captions
are both diverse and high quality.
Table~\ref{tab:diversity} shows some samples when returning the N-best list from our
beam search decoder instead of the best hypothesis. Notice how the samples are
diverse and may show different aspects from the same image.
The agreement in BLEU score between the top 15 generated sentences is 58, which is similar to that of humans among them. This indicates the amount of diversity
our model generates.
In bold are the sentences that
are not present in the training set. If we take the best candidate, the
sentence is present in the training set 80\% of the times.
This is not too surprising given that the amount
of training data is quite small, so it is relatively easy for the model to pick ``exemplar''
sentences and use them to generate descriptions.
If we instead analyze the top 15 generated sentences, about half of the times we
see a completely novel description, but still with a similar BLEU score,
indicating that they are of enough quality, yet they
provide a healthy diversity.

\begin{table}[htb]
\caption{{N-best examples from the MSCOCO test set. Bold lines indicate a novel sentence not present in the training set.}}
\label{tab:diversity}
\begin{center}
\begin{tabular}{|l|}\hline
A man throwing a frisbee in a park. \\
{\bf A man holding a frisbee in his hand.} \\
{\bf A man standing in the grass with a frisbee.} \\
\hline
A close up of a sandwich on a plate. \\
A close up of a plate of food with french fries. \\
A white plate topped with a cut in half sandwich. \\
\hline
A display case filled with lots of donuts. \\
{\bf A display case filled with lots of cakes.} \\
{\bf A bakery display case filled with lots of donuts.} \\
\hline
\end{tabular}
\end{center}
\end{table}

\subsubsection{Ranking Results}

While we think ranking is an unsatisfactory way to evaluate description
generation from images, many papers report ranking scores,
using the set of testing captions as candidates to rank given a test image.
The approach that works best on these metrics (MNLM),
specifically implemented a ranking-aware loss. Nevertheless,
NIC is doing surprisingly well on both ranking tasks (ranking descriptions
given images, and ranking images given descriptions),
as can be seen in
Tables~\ref{tab:recall@10} and~\ref{tab:recall@1030k}. Note that for the Image Annotation task, we normalized our scores similar to what~\cite{baidu2014} used.

\begin{table}
\caption{Recall@k and median rank on Flickr8k.\label{tab:recall@10}}
\centering
\begin{small}
\setlength{\tabcolsep}{3pt}
\begin{tabular}{|c|ccc|ccc|}
 \hline
\multirow{2}{*}{Approach} & \multicolumn{3}{c|}{Image Annotation} & \multicolumn{3}{c|}{Image Search} \\
 & R@1 & R@10 & Med $r$ &  R@1 & R@10 & Med $r$ \\
\hline
\hline
DeFrag~\cite{karpathy2014deep} & 13 & 44 & 14             &    10 & 43 & 15  \\
m-RNN~\cite{baidu2014}         &  15 & 49 & 11               &  12 & 42 & 15\\
MNLM~\cite{kiros2014}        &  18   & 55 & 8        &  13 & 52 & 10   \\
\hline
NIC                            &  \bf{20} & \bf{61} & \bf{6}              &    \bf{19} & \bf{64} & \bf{5} \\
\hline
\end{tabular}
\end{small}
\end{table}

\begin{table}
\caption{Recall@k and median rank on Flickr30k.\label{tab:recall@1030k}}
\centering
\begin{small}
\setlength{\tabcolsep}{3pt}
\begin{tabular}{|c|ccc|ccc|}
\hline
\multirow{2}{*}{Approach} & \multicolumn{3}{c|}{Image Annotation} & \multicolumn{3}{c|}{Image Search} \\
 & R@1 & R@10 & Med $r$ &  R@1 & R@10 & Med $r$ \\
\hline
\hline
DeFrag~\cite{karpathy2014deep} & 16 & 55 & 8             &    10 & 45 & 13  \\
m-RNN~\cite{baidu2014}         &  18 & 51 & 10               &  13 & 42 & 16\\
MNLM~\cite{kiros2014}        &  \bf{23}   & \bf{63} & \bf{5}        &  \bf{17} & \bf{57} & \bf{8}   \\
\hline
NIC                            &  17 & 56  & 7               &    \bf{17} & \bf{57} & \bf{7} \\
\hline
\end{tabular}
\end{small}
\end{table}

\subsubsection{Human Evaluation}
\label{sec:human}

\begin{figure}
\begin{center}
  \includegraphics[width=1.0\columnwidth]{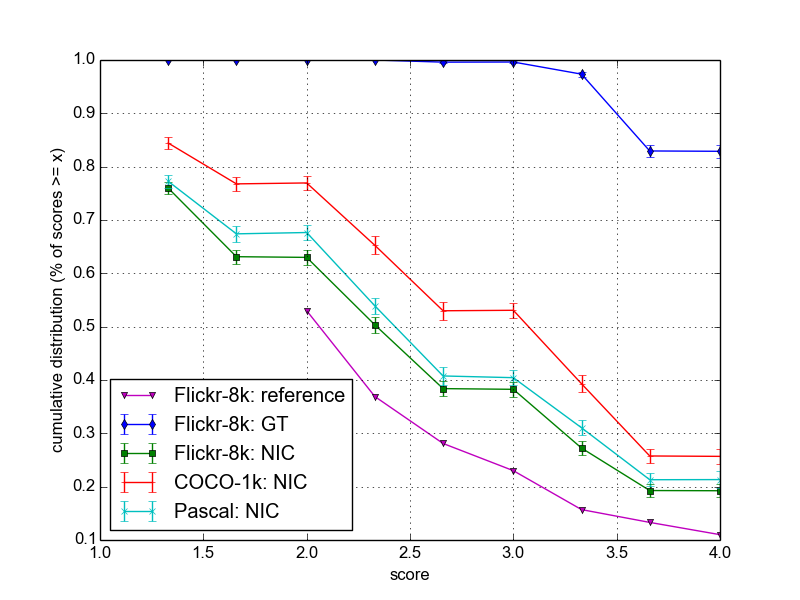}
\end{center}
\vspace{-0.5cm}
\caption{\label{fig:turk_eval_numeric} {\em Flickr-8k: NIC}: predictions produced by NIC on the Flickr8k test set (average score: 2.37); {\em Pascal: NIC}: (average score: 2.45); {\em COCO-1k: NIC}: A subset of 1000 images from the MSCOCO test set with descriptions produced by NIC (average score: 2.72); {\em Flickr-8k: ref}: these are results from~\cite{hodosh2013framing} on Flickr8k rated using the same protocol, as a baseline (average score: 2.08); {\em Flickr-8k: GT}: we rated the groundtruth labels from Flickr8k using the same protocol. This provides us with a ``calibration'' of the scores (average score: 3.89)}
\end{figure}

\begin{figure*}
\begin{center}
  \includegraphics[width=\textwidth]{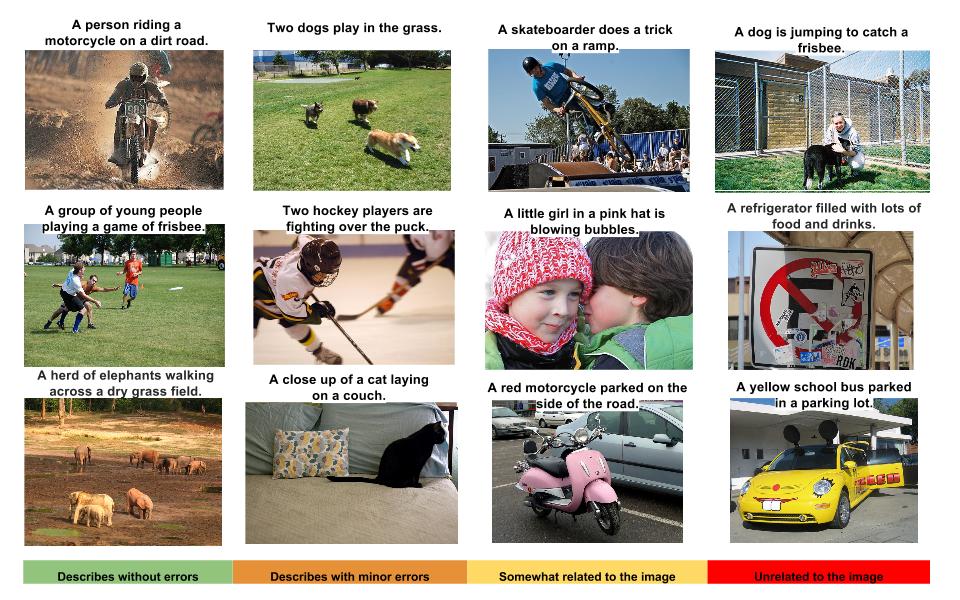}
\vspace{-1cm}
\end{center}
\caption{\label{fig:turk_eval_examples} A selection of evaluation results, grouped by human rating.}
\end{figure*}

Figure~\ref{fig:turk_eval_numeric} shows the result of the human evaluations
of the descriptions provided by NIC, as well as a reference system and
groundtruth on various datasets. We can see that NIC is better than the reference
system, but clearly worse than the groundtruth, as expected.
This shows that BLEU is not a perfect metric, as it does not capture well
the difference between NIC and human descriptions assessed by raters.
Examples of rated images can be seen in Figure~\ref{fig:turk_eval_examples}.
It is interesting to see, for instance in the second image of the first
column, how the model was able to notice the frisbee given its size.

\subsubsection{Analysis of Embeddings}

In order to represent the previous word $S_{t-1}$ as input to the decoding LSTM
producing $S_t$, we use word embedding vectors~\cite{mikolov2013},
which have the advantage of
being independent of the size of the dictionary (contrary to a simpler
one-hot-encoding approach).
Furthermore, these word embeddings can be jointly trained with the rest of the
model. It is remarkable to see how the learned representations
have captured some semantic from the statistics of the language.
Table~\ref{tab:embeddings} shows, for a few example words, the nearest other
words found in the learned embedding space.

Note how some of the relationships
learned by the model will help the vision component. Indeed, having ``horse'', ``pony'',
and ``donkey'' close to each other will encourage the CNN to extract features that
are relevant to horse-looking animals.
We hypothesize that, in the extreme case where we see very few examples of a class (e.g., ``unicorn''),
its proximity to other word embeddings (e.g., ``horse'') should
provide a lot more information that would be completely lost with more
traditional bag-of-words based approaches.

\begin{table}[htb]
\caption{{Nearest neighbors of a few example words}}
\label{tab:embeddings}
\begin{center}
\begin{tabular}{|l|l|}\hline
Word & Neighbors \\ \hline
car & van, cab, suv, vehicule, jeep \\
boy & toddler, gentleman, daughter, son \\
street & road, streets, highway, freeway \\
horse & pony, donkey, pig, goat, mule \\
computer & computers, pc, crt, chip, compute \\ \hline
\end{tabular}
\end{center}
\end{table}

\section{The MS COCO Image Captioning Challenge}
\label{sec:competition}

In the spring of 2015, as part of the MS COCO dataset a challenge was organized\footnote{More details can be found+
on the competition website: \url{http://mscoco.org/dataset/#captions-challenge2015}.}.
Participants were recommended to train their algorithms on the MS COCO 2014
dataset, and results on the validation and test sets were submitted on an
evaluation server, with no more than 5 attempts in total per group, in order
to limit overfitting on the test set.
Human judges then evaluated the competing approaches
and the winners were invited to present their approach at a workshop organized
during CVPR 2015.

We entered the competition and the rest of this section
explains the various techniques we have explored in this context, building
on our baseline model described in the previous sections.

\subsection{Metrics}

The metrics used are discussed in in Section~\ref{sec:exps}. A special emphasis is on CIDER~\cite{cider}, which was 
chosen by the competition organizers to rank teams. As a result we use also during hyper-parameter selection.

\begin{table}
\caption{Pearson correlation and human rankings found in the MSCOCO official website competition table for several automatic metrics (using 40 ground truth captions in the test set).}
\label{tab:correl}
\begin{center}
\begin{tabular}{|l|rr|}
          \hline
          & Correlation  (vs CIDER)  & Human Rank \\
          \hline
CIDER & 1.0  & 6 \\
METEOR & 0.98 & 3 \\
ROUGE & 0.91 & 11 \\
BLEU-4 & 0.87 & 13 \\
\hline
\end{tabular}
\end{center}
\end{table}

We found all the automatic metrics to correlate
with each other quite strongly (see Table~\ref{tab:correl}). Notably, the main difference of these metrics is on how humans rank on it versus several automatic image captioning systems (such as the one we propose).
Interestingly, BLEU score seems to be quite bad (humans rank 13th out of 16); CIDER fares better
(where humans rank 6th); METEOR is the automatic metric where humans rank the highest
(third).

\subsection{Improvements Over Our CVPR15 Model}
\label{sec:improvements}
In this Section we analyze what components were improved with respect to the model which we originally
studied in our CVPR 2015 work \cite{google2014}. Section~\ref{sec:comp_resutls} shows a summary of
the results on both automatic and human metrics from the MSCOCO competition. We summarize all the
improvements in Table~\ref{tab:improvements}. For reproducibility, we also open source an implementation of our model in TensorFlow \cite{TFPAPER}\footnote{\url{https://github.com/tensorflow/models/tree/master/im2txt}}.

\begin{table}
\caption{A summary of all the improvements which we introduced for the MSCOCO competition. The reported improvements are on BLEU-4, but similar improvements are consistent across all the metrics.}
\label{tab:improvements}
\begin{center}
\begin{tabular}{|l|r|}
          \hline
Technique& BLEU-4 Improvement \\
          \hline
Better Image Model~\cite{batchnorm} & 2 \\
Beam Size Reduction & 2 \\
Fine-tuning Image Model & 1 \\
Scheduled Sampling~\cite{scheduled_sampling} & 1.5 \\
Ensembles & 1.5 \\
\hline
\end{tabular}
\end{center}
\end{table}

\subsubsection{Image Model Improvement}

When we first submitted our image captioning paper to CVPR 2015, we used
the best convolutional neural network at the time, known as
GoogleLeNet~\cite{szegedy2014going}, which had 22 layers, and was the winner
of the 2014 ImageNet competition. Later on, an even better approach was proposed
in~\cite{batchnorm} and included a new method, called {\em Batch Normalization},
to better normalize each layer of a neural network with respect to the current
batch of examples, so as to be more robust to nonlinearities. The new
approach got significant improvement on the ImageNet task (going from 6.67\%
down to 4.8\% top-5 error) and the MSCOCO image captioning task, improving
BLEU-4 by 2 points absolute.

\subsubsection{Image Model Fine Tuning}
In the original set of experiments, to avoid overfitting we initialized the image convolutional
network with a pre-trained model (we first used GoogleLeNet, then switched
to the better Batch Normalization model), but then fixed its parameters
and only trained the LSTM part of the model on the MS COCO training set.

For the competition, we also considered adding some fine
tuning of the image model while training the LSTM, which helped the image
model focus more on the kind of images provided in the MS COCO training set,
and ended up improving the performance on the captioning task.

It is important to note that fine tuning the image model must be carried after the
LSTM parameters have settled on a good language model: we found that, when
jointly training both, the noise in the initial gradients coming from the LSTM into
the image model corrupted the CNN and would never recover. Instead, we train
for about 500K steps (freezing the CNN parameters), and then switch to jointly
train the model for an additional 100K steps. Training was done using a single GPU (Nvidia K20), and step time was about 3 seconds. Thus, training took over 3 weeks -- parallelizing training yielded somewhat worse results, though it increased the speed to convergence.

The improvements achieved by this was 1 BLEU-4 point. More importantly, this
change allowed the model to transfer information from the image to the language
which was likely not possible due to the insufficient coverage of the ImageNet
label space. For instance, after the change we found many examples where we
predict the right colors, e.g. ``A blue and yellow train ...''.  It is
plausible that the top-layer CNN activations are overtrained on
ImageNet-specific classes and could throw away interesting features (such as
color), thus the caption generation model may not output words corresponding
to those features, without fine tuning the image model.

\subsubsection{Scheduled Sampling}

As explained in Section~\ref{sec:lstm}, our model uses an LSTM to generate
the description given the image. As shown in Figure~\ref{fig:unrolled_lstm},
LSTMs are trained by trying to predict each word of the caption given the
current state of the model and {\bf the previous word} in the caption.
At inference, for a new image, the previous word is obviously unknown and
is thus replaced by the word generated by the model itself at the previous
step. There is thus a discrepancy between training and inference. Recently,
we proposed~\cite{scheduled_sampling} a curriculum learning strategy to
gently change the training process from a fully guided scheme using the
true previous word, towards a less guided scheme which mostly uses the
model generated word instead. We applied this strategy using various
schedules for the competition, and found that it improved up to 1.5 BLEU-4 points
over using the standard training objective function.

\subsubsection{Ensembling}

Ensembles~\cite{bagging} have long been known to be a very simple yet effective
way to improve performance of machine learning systems. In the context of
deep architectures, one only needs to train separately multiple models on
the same task, potentially varying some of the training conditions,
and aggregating their answers at inference time. For the competition, we
created an ensemble of 5 models trained with {\em Scheduled Sampling} and
10 models trained with fine-tuning the image model. The resulting
model was submitted to the competition, and it further improved our results
by 1.5 BLEU-4 points.

\subsubsection{Beam Size Reduction}
\label{sec:beam_size_reduction}

In order to generate a sentence with our proposed approach, we described
in Section~\ref{sec:lstm} the use of BeamSearch, where we maintain a list
of the top-$k$ sequences of words generated so far. In the original paper,
we tried only two values for $k$: 1 (which means only keep the best generated
word according to the model at each time step) and 20.

For the competition, we actually tried several more beam sizes, and selected
the size which generated the best sequences of words according to the
CIDER metric, which we consider to be the metric most aligned with human
judgements. Contrary to our expectations, the best beam size turned out to
be small: 3.

Note that, as the beam size increases, we score more candidate
sentences and pick the best according to the obtained likelihood. Hence,
if the model was well trained and the likelihood was aligned with human
judgement, increasing the beam size should always yield better sentences.
The fact that we obtained the best performance with a relatively small beam
size is an indication that either the model has overfitted or the objective
function used to train it (likelihood) is not aligned with human judgement.

We also observed that, by reducing the beam size (i.e., with a shallower search
over sentences), we increase the novelty of generated sentences. Indeed,
instead of generating captions which repeat training captions 80\% of the time,
this gets reduced to 60\%. This hypothesis supports the fact that the model
has overfitted to the training set, and we see this reduced beam size technique
as another way to regularize (by adding some noise to the inference process).

Reducing the beam size was the single change that improved our CIDER score
the most. This simple change yielded more than 2 BLEU-4 points improvement.

\begin{figure*}
\begin{center}
\includegraphics[width=\textwidth]{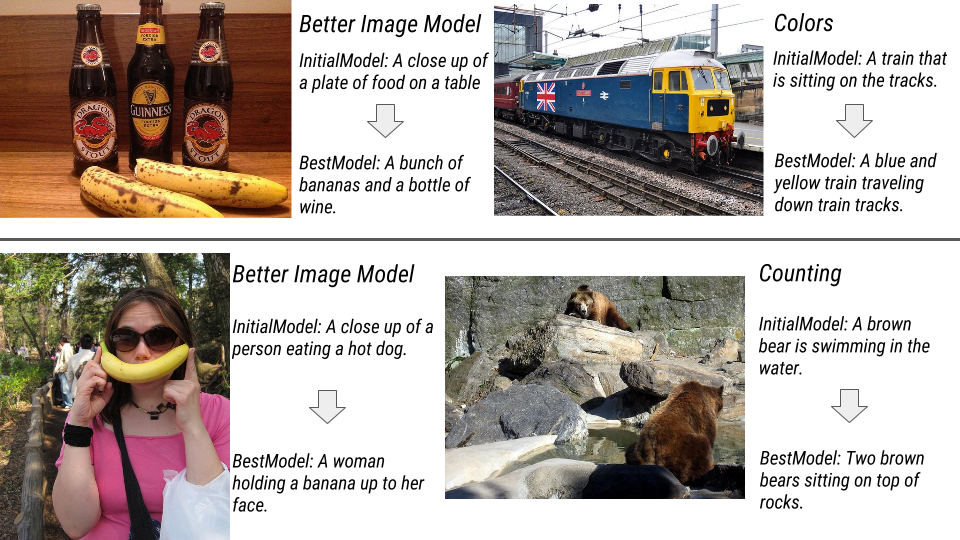}
\end{center}
\caption{\label{fig:examples} A selection of evaluation images, comparing
the captions obtained by our original model (InitialModel) and the model submitted to the
competition (BestModel).}
\end{figure*}

\subsection{Competition Results}
\label{sec:comp_resutls}
\subsubsection{Automatic Evaluation}
All the teams were allowed up to 5 submissions to the evaluation server on a large, unseen
set of test images. The leaderboard allowed for teams to monitor progress, and it motivated
us to keep improving the accuracy of our model up to the deadline. Despite the automatic
metrics not fully characterizing the quality of the captions, strong correlations were present
(i.e., improving an automatic metric generally implied a better captioning system).

Since we submitted our paper, and thanks to all the improvements, our BLEU-4 score
improved by 8 points absolute (see Section~\ref{sec:improvements}). The top 5 submission
according to the automatic metrics on the test set (sorted by CIDER, and using 5 ground truth captions) are presented 
in Table~\ref{table:top5_aut}:

\begin{footnotesize}
\setlength{\tabcolsep}{0.5em}
\begin{table}[h]
\caption{\label{table:top5_aut} Automatic scores of the top five competition submissions.}
\begin{tabular}{|l|rrrr|l|}
\hline
          & CIDER    & METEOR    & ROUGE    & BLEU-4  & Rank \\
\hline
Google \cite{google2014}	&0.943&0.254&	0.53&	0.309& 1st \\
MSR Captivator \cite{devlin2015language}&	0.931&	0.248&	0.526	&0.308& 2nd\\
m-RNN \cite{mao2014deep}&	0.917	&0.242&	0.521&	0.299& 3rd \\
MSR \cite{msr14} 	&0.912	&0.247	&0.519	&0.291& 4th \\
m-RNN (2) \cite{mao2014deep}&	0.886	&0.238	&0.524&	0.302& 5th\\
\hline
Human& 0.854	&0.252&	0.484	&0.217 & 8th \\
\hline
\end{tabular}
\vspace{2pt}
\end{table}
\end{footnotesize}

\subsubsection{Human Evaluation}

The most promising 15 submissions to the MSCOCO challenge, as well as a
human baseline, were evaluated on 5 different metrics:
\begin{description}
\item[M1] Percentage of captions that are evaluated as better or equal to human caption.
\item[M2] Percentage of captions that pass the Turing Test.
\item[M3] Average correctness of the captions on a scale 1-5 (incorrect - correct).
\item[M4] Average amount of detail of the captions on a scale 1-5 (lack of details - very detailed).
\item[M5] Percentage of captions that are similar to human description.
\end{description}

Note that M1 and M2 were the ones used to decide the winner. The others were merely experimental, but are reported here for completeness.

Results are available on the Leaderboard of the competition website at \url{http://mscoco.org/dataset/#captions-leaderboard}.
The top 5 submissions according to these metrics (sorted by M1+M2) are shown in Table~\ref{table:top5_human}:

\begin{footnotesize}
\setlength{\tabcolsep}{0.5em}
\begin{table}[h]
\caption{\label{table:top5_human} Human generated scores of the top five competition submissions.}
\begin{tabular}{|l|rrrrr|l|}
\hline
          & M1    & M2    & M3    & M4    & M5    & Rank \\
\hline
Google \cite{google2014}   & 0.273 & 0.317 & 4.107 & 2.742 & 0.233 & 1st \\
MSR \cite{msr14}       & 0.268 & 0.322 & 4.137 & 2.662 & 0.234 & 1st \\
MSR Captivator \cite{devlin2015language} & 0.250 & 0.301 & 4.149 & 2.565 & 0.233 & 3rd \\
Montreal/Toronto \cite{xu2015show} & 0.262 & 0.272 & 3.932 & 2.832 & 0.197 & 3rd \\
Berkeley LRCN \cite{berkeley2014} & 0.246 & 0.268 & 3.924 & 2.786 & 0.204 & 5th \\
\hline
Human & 0.638	 & 0.675 &	4.836&	3.428&	0.352 & 1st \\
\hline
\end{tabular}
\end{table}
\end{footnotesize}

\vspace{10pt}
Finally, we show in Figure~\ref{fig:examples} a few example images together
with the caption obtained by our original model, compared with the caption
obtained by the final model submitted to the competition. We took a random
sample of 20 images from the development set, and picked the ones that looked
most interesting (all of them had a better caption except for one). It is clear that
the overall quality of the captions have improved significantly, a fact that should be
obvious given the overall improvement in BLEU-4 from the improvements that
we showed in this section was 8 points absolute.

\section{Conclusion}
\label{sec:conclusion}
We have presented NIC, an
end-to-end neural network system that can automatically view an image
and generate a reasonable description in plain English.
NIC is based on a convolution neural network that encodes an image into
a compact representation, followed by a recurrent neural network that
generates a corresponding sentence. The model is trained to maximize
the likelihood of the sentence given the image.
Experiments on several datasets
show the robustness of NIC in terms of qualitative results (the
generated sentences are very reasonable) and quantitative evaluations,
using either ranking metrics or BLEU, a metric used in machine translation
to evaluate the quality of generated sentences.
Based on our initial results, we participated in the 2015 MS COCO challenge
comparing approaches on the task of image captioning. We presented and
analyzed in this paper the various improvements we have made to our basic
NIC model and described the competition results which ranked our model in
first position using both automatic and human evaluations.
It is clear from these experiments that, as the size of the available
datasets for image description increases, so will the performance of
approaches like NIC.

Despite the exciting results on captioning, we believe it is just the beginning. The 
produced descriptions are one of many possible image interpretations. One possible
direction is the have a system which is capable of more targeted descriptions -- either 
anchoring the descriptions to given image properties and locations or being a response
to a user specified question or task. Further research direction are better evaluation
metrics or evaluation through higher level goals found in application such as robotics.



%



\ifCLASSOPTIONcompsoc
  \section*{Acknowledgments}
\else
  \section*{Acknowledgment}
\fi

We would like to thank Geoffrey Hinton, Ilya Sutskever, Quoc Le, Vincent Vanhoucke,
and Jeff Dean for useful discussions on the ideas behind the paper, and the
write up. Also many thanks to Chris Shallue for driving the efforts to reimplement and open source our model in TensorFlow.

\ifCLASSOPTIONcaptionsoff
  \newpage
\fi



\bibliographystyle{IEEEtran}
\bibliography{egbib}
%
%
%

%

\begin{IEEEbiography}[{\includegraphics[width=1in,height=1.25in,clip,keepaspectratio]{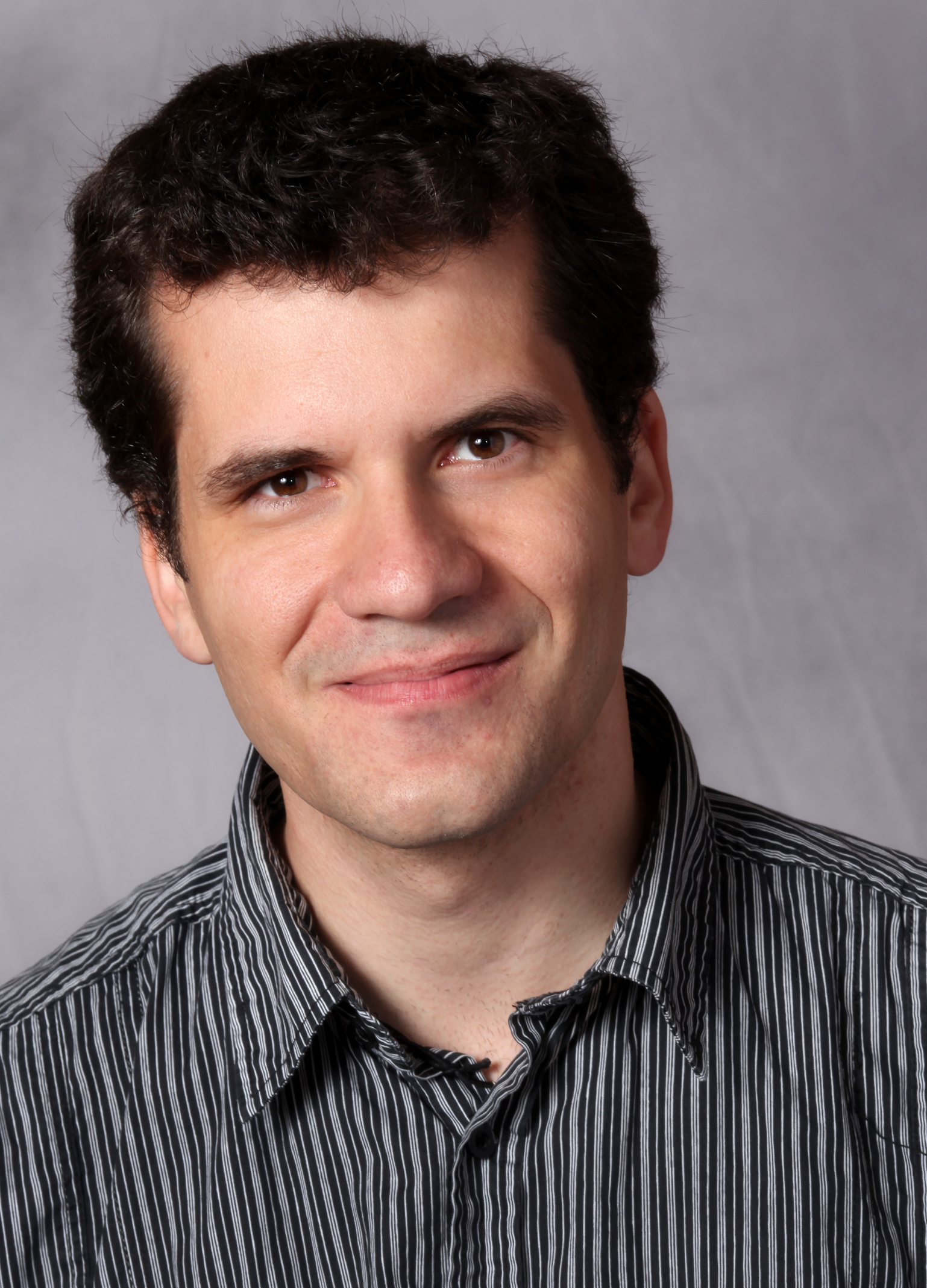}}]{Oriol Vinyals} is a Research Scientist at Google DeepMind, working in Deep Learning. Prior to joining DeepMind, Oriol was part of the Google Brain team. He holds a Ph.D. in EECS from University of California, Berkeley, and a Masters degree from University of California, San Diego. He is a recipient of the 2011 Microsoft Research PhD Fellowship. He was an early adopter of the new deep learning wave at Berkeley, and in his thesis he focused on non-convex optimization and recurrent neural networks. At Google DeepMind he continues working on his areas of interest, which include artificial intelligence, with particular emphasis on machine learning, language, and vision.
\end{IEEEbiography}

\begin{IEEEbiography}[{\includegraphics[width=1in,height=1.25in,clip,keepaspectratio]{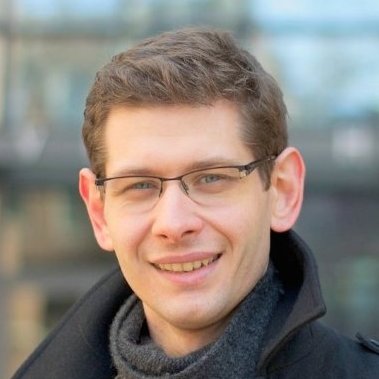}}]{Alexander Toshev} obtained a Diplom in Computer Science from University of Karlsruhe in 2005, a PhD in Computer and Information Sciences from University of Pennsylvania in 2010. His doctoral work was awarded the 2011 Morris and Dorothy Rubinoff Award by the Engineering Faculty. Since 2011 he has been a research scientist at Google. His research interests lie broadly in computer vision and machine learning, in particular image understanding (object detection, human pose estimation, image segmentation) and synergy of text and images. \end{IEEEbiography}


\begin{IEEEbiography}[{\includegraphics[width=1in,height=1.25in,clip,keepaspectratio]{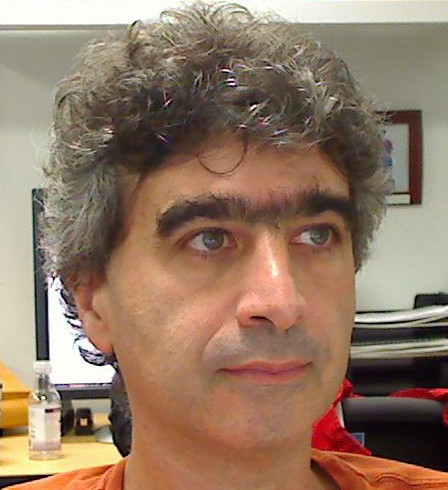}}]{Samy Bengio}
(PhD in computer science, University of Montreal, 1993) is a research scientist at Google since 2007. Before that, he was senior researcher in statistical machine learning at IDIAP Research Institute since 1999. His most recent research interests are in machine learning, in particular deep learning, large scale online learning, image ranking and annotation, music and speech processing. He is action editor of the Journal of Machine Learning Research and on the editorial board of the Machine Learning Journal. He was associate editor of the journal of computational statistics, general chair of the Workshops on Machine Learning for Multimodal Interactions (MLMI'2004-2006), programme chair of the International Conference on Learning Representations (ICLR'2015-2016), programme chair of the IEEE Workshop on Neural Networks for Signal Processing (NNSP'2002), chair of BayLearn (2012-2015), and several times on the programme committee of international conferences such as NIPS, ICML, ECML and ICLR. More information can be found on his website: http://bengio.abracadoudou.com.
\end{IEEEbiography}

\begin{IEEEbiography}[{\includegraphics[width=1in,height=1.25in,clip,keepaspectratio]{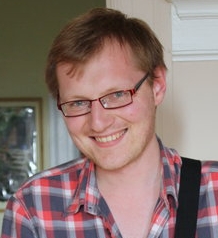}}]{Dumitru Erhan}
 (PhD in computer science, University of
  Montreal, 2011) is a software engineer at Google since 2012. Before that, he
  was scientist at Yahoo! Labs from 2011 to 2012. His research interests span
  the intersection of deep learning, computer vision and natural language.  In
  particular, he is interested in efficient models for understanding what and
  where is in image, as well as for answering arbitrary questions about them.
  More information at http://dumitru.ca.  \end{IEEEbiography}




\end{document}